\documentclass[sigconf,screen,nonacm]{acmart}

\setcopyright{none}
\copyrightyear{}
\acmYear{}
\acmDOI{}
\acmPrice{}
\acmISBN{}

\usepackage{subfigure}
\usepackage{float}
\usepackage{graphicx}

\definecolor{groupgreen}{HTML}{A8D4AD}
\definecolor{groupred}{HTML}{967D69}
\definecolor{groupblue}{HTML}{92B9BD}
\definecolor{infoyellow}{HTML}{F2F79E}
\definecolor{datayellow}{HTML}{E8EC67}

\definecolor{color_corr_skip}{HTML}{1b998b} 
\definecolor{color_not_corr_skip}{HTML}{e71d36} 
\definecolor{color_info}{HTML}{c5d86d} 
\definecolor{color_state}{HTML}{2e294e} 

\usepackage{subcaption}
\usepackage{bm}
\usepackage{pifont}
\usepackage{multirow}
\usepackage{enumitem}

\newif\ifshowrelnodes
\showrelnodesfalse
\begin{document}
\newcommand{\cmark}{\ding{51}}
\newcommand{\xmark}{\ding{55}}
\newcommand{\singleasterisk}[1]{\node[above=-0.125cm of #1,anchor=west] {$^{*}$};}
\newcommand{\doubleasterisk}[1]{\node[above=-0.125cm of #1,anchor=west] {$^{**}$};}
\title{Exploring Student Expectations and Confidence in Learning Analytics}
\author{Hayk Asatryan}
\orcid{0000-0001-5631-4779}
\email{hayk.asatryan@hs-bochum.de}
\affiliation{%
  \institution{Bochum University of Applied Sciences}
  \streetaddress{Kettwiger Stra\ss e 20}
  \city{Heiligenhaus}
  \state{North Rhine-Westphalia}
  \country{Germany}
  \postcode{42579}
}

\author{Basile Tousside}
\orcid{0000-0002-9332-5060}
\email{basile.tousside@hs-bochum.de}
\affiliation{%
  \institution{Bochum University of Applied Sciences}
  \streetaddress{Kettwiger Stra\ss e 20}
  \city{Heiligenhaus}
  \state{North Rhine-Westphalia}
  \country{Germany}
  \postcode{42579}
}

\author{Janis Mohr}
\orcid{0000-0001-6450-074X}
\email{janis.mohr@hs-bochum.de}
\affiliation{%
  \institution{Bochum University of Applied Sciences}
  \streetaddress{Kettwiger Stra\ss e 20}
  \city{Heiligenhaus}
  \state{North Rhine-Westphalia}
  \country{Germany}
  \postcode{42579}
}

\author{Malte Neugebauer}
\orcid{0000-0002-1565-8222}
\email{malte.neugebauer@hs-bochum.de}
\affiliation{%
  \institution{Bochum University of Applied Sciences}
  \streetaddress{Kettwiger Stra\ss e 20}
  \city{Heiligenhaus}
  \state{North Rhine-Westphalia}
  \country{Germany}
  \postcode{42579}
}

\author{Hildo Bijl}
\orcid{0000-0002-7021-7120}
\email{hildo.bijl@hs-bochum.de}
\affiliation{%
  \institution{Bochum University of Applied Sciences}
  \streetaddress{Gesundheitscampus 6 – 8}
  \city{Bochum}
  \state{North Rhine-Westphalia}
  \country{Germany}
  \postcode{44801}
}

\author{Paul Spiegelberg}
\orcid{0009-0004-9528-7143}
\email{paul.spiegelberg@hs-gesundheit.de}
\affiliation{%
  \institution{Hochschule für Gesundheit}
  \streetaddress{Gesundheitscampus 6 – 8}
  \city{Bochum}
  \state{North Rhine-Westphalia}
  \country{Germany}
  \postcode{44801}
}

\author{Claudia Frohn-Schauf}
\orcid{0009-0005-4311-7051}
\email{claudia.frohn-schauf@hs-bochum.de}
\affiliation{%
  \institution{Bochum University of Applied Sciences}
  \streetaddress{Am Hochschulcampus 1}
  \city{Bochum}
  \state{North Rhine-Westphalia}
  \country{Germany}
  \postcode{44801}
}

\author{J\"org Frochte}
\orcid{0000-0002-5908-5649}
\email{joerg.frochte@hs-bochum.de}
\affiliation{%
  \institution{Bochum University of Applied Sciences}
  \streetaddress{Kettwiger Stra\ss e 20}
  \city{Heiligenhaus}
  \state{North Rhine-Westphalia}
  \country{Germany}
  \postcode{42579}
}
\begin{abstract}
Learning Analytics (LA) is nowadays ubiquitous in many educational systems, providing the ability to collect and analyze student data in order to understand and optimize learning and the environments in which it occurs. On the other hand, the collection of data requires to comply with the growing demand regarding privacy legislation. In this paper, we use the Student Expectation of Learning Analytics Questionnaire (SELAQ) to analyze the expectations and confidence of students from different faculties regarding the processing of their data for Learning Analytics purposes. This allows us to identify four clusters of students through clustering algorithms: {\itshape Enthusiasts}, {\itshape Realists}, {\itshape Cautious} and {\itshape Indifferents}. This structured analysis provides valuable insights into the acceptance and criticism of Learning Analytics among students. 
\end{abstract}
\begin{CCSXML}
<ccs2012>
   <concept>
       <concept_id>10010405.10010489.10010495</concept_id>
       <concept_desc>Applied computing~E-learning</concept_desc>
       <concept_significance>500</concept_significance>
       </concept>
   <concept>
       <concept_id>10010147.10010257.10010258.10010260.10003697</concept_id>
       <concept_desc>Computing methodologies~Cluster analysis</concept_desc>
       <concept_significance>500</concept_significance>
       </concept>
   <concept>
       <concept_id>10003456.10003457.10003527.10003542</concept_id>
       <concept_desc>Social and professional topics~Adult education</concept_desc>
       <concept_significance>500</concept_significance>
       </concept>
 </ccs2012>
\end{CCSXML}

\ccsdesc[500]{Applied computing~E-learning}
\ccsdesc[500]{Computing methodologies~Cluster analysis}
\ccsdesc[500]{Social and professional topics~Adult education}
\keywords{Learning Analytics, Survey, Data Protection, Clustering}
\maketitle
\section{Introduction}
\label{sec:Introduction}
Recently, Learning Analytics (LA) has emerged as an appealing tool in optimizing learning dynamics within educational systems. It is the process of collecting, analyzing and interpreting data about learners and their contexts \cite{Picciano2012}. The purpose is to understand and optimize learning, thus offering valuable insights into the dynamic interplay between learning environments and student performance. An example of the use of LA  provides the Open University, which has implemented initiatives to improve retention rates ~\cite{calvert2014developing}. However, collecting and analyzing student data raise significant concerns regarding Data Protection (DP) ~\cite{ifenthaler2016student, Rodda2022}. At the same time, DP laws, legislations and regulations (e.g., GDPR) are increasingly tightened to safeguard the personal data of individuals ~\cite{Karunaratne2021}. The term {\itshape Data Protection} is used herein to encompass both data security and individual privacy concerns. As a consequence, balancing the increasing demands on Data Protection \& Privacy  and making use of educational big data has become a mandatory challenge. Fostering a deep understanding of students' acceptance regarding data processing for LA is therefore paramount in shaping effective and ethical LA practices. 

To address this issue, the current paper delves into the acceptance of LA among students by empirically scrutinizing their attitudes. Without active involvement and support from students, such a  system cannot be effectively utilized. Students might disapprove or actively avoid using the offered resources if they feel uncomfortable with the system. While educators also need to see tangible benefits, especially as the introduction of new systems and tools is linked to additional work, our primary focus in this paper is on understanding the student perspective. Specifically, we analyze the expectations and confidence levels held by students from various degree courses concerning the processing of their data for Learning Analytics purposes. 

To achieve this, we conduct an anonymous survey among students using an adapted version of the well-known $12$-item Student Expectations of Learning Analytics Questionnaire (SELAQ), tool for collecting quantitative measures of students’ expectations of LA services ~\cite{whitelock2019student}. Student expectations of LA are effectively measured through the integration of two distinct subscales within the questionnaire: (i) desire and (ii) expectation. In order to cluster students according to their varying needs and desires, we apply machine learning techniques to the survey results. This approach aims to uncover not only clusters based on discernible characteristics like academic disciplines, but also to identify nuanced groups that might not be immediately apparent through traditional metrics. As a result, we identify four primary student clusters: {\itshape Enthusiasts}, {\itshape Realists}, {\itshape Cautious} and {\itshape Indifferents}. 

A more in-depth exploration of these clusters and their defining attributes is presented in Section \ref{sec:Study:Analysing:the:Questionaire:Results}. 
{\itshape Enthusiasts} and {\itshape Realists} share comparably high values with respect to the desire for DP and LA implementation. However, they differ in their expectations regarding the practical achievement of these aspirations. In contrast, cautious students hold reservations about the 
effectiveness of Learning Analytics and might necessitate greater persuasion and reassurance concerning the implementation of such systems. Indifferent students on the other hand, exhibit a pervasive lack of interest or involvement. Aiming to detect prevailing trends, we further analyze these clusters with respect to students' specializations. Notably, the results reveal that students specializing in Sustainability tend to exhibit a high level of enthusiasm toward the idea of LA, whereas those pursuing Architecture display a more pronounced skepticism toward its implementation.
Our main contributions can be summarized as follows:
\begin{itemize}
    \item We present the results of a comprehensive survey conducted among university students. This survey sheds light on their attitudes and expectations concerning LA systems.
    \item We examine the survey results and distinguishing responses among student groups, classified by their academic disciplines.
    \item We employ both data analysis and an automatic clustering approach to effectively segment the student population, which offers insightful perspectives on the interpretation of the student dynamics.
    \item Building upon the insights derived from this clustering, we implement a decision-based approach to enhance the clusters transparency and explainability. This enables a deeper understanding of student behaviors and preferences.
    \item Finally, we address notable challenges and share the lessons learned regarding student acceptance toward Learning Analytics systems. These insights lay the foundation for successful and broadly accepted LA implementations at universities.
\end{itemize}

\section{Related Work}
\label{sec:Previous:Work}
Learning Analytics has emerged as a vital tool for enhancing educational quality \cite{leitner2017learning, viberg2018current}.  
Its success hinges on student acceptance. Numerous studies probe factors influencing student attitudes towards LA, highlighting the importance of student engagement and their perspectives \cite{hantoobi2021review,viberg2022exploring}.
For instance, \cite{schumacher2018features} explores students' expectations of LA system features, emphasizing tools aiding in planning, self-assessments, adaptive recommendations and personalized learning activity analyses. 
\cite{ifenthaler2016student} drives a privacy-focused approach and analyzes student perceptions of privacy principles related to LA. The results indicate that students expect Learning Analytics systems to include adaptive and personalized dashboards. More recently, \cite{mutimukwe2022students} explored students' privacy concerns regarding LA practice in higher education and found that students' perceptions of privacy control and privacy risks determine their trusting beliefs. 
In~\cite{Uskov2018}, concerns about anonymized data presentation and prediction algorithm quality were raised. In the past few years, the Student Expectation of Learning Analytics Questionnaire (SELAQ) has emerged as a central tool in LA studies. Introduced in 2019 by Whitelock et al. \cite{whitelock2019student}, it has established itself as a fundamental resource in capturing student perceptions, desires and expectations in the field of Learning Analytics  \cite{whitelock2020assessing,WhitelockWainwright2021, Wollny2023}. For instance, in \cite{Wollny2023}, SELAQ was used to survey $417$ European students, revealing a generally positive student attitude towards LA. Similarly, \cite{pontual2022penny} uses SELAQ to investigate Brazilian students' expectations of LA. The study's findings indicated that the surveyed students held high expectations for the implementation of LA in their institutions, corroborating earlier research from Europe \cite{kollom2021four} and Latin America \cite{garcia2021aligning}.
Prior studies using SELAQ have not segregated students based on personal or university traits. However, recognizing discipline-specific nuances could offer a more holistic understanding of student LA attitudes. The present paper considers students' disciplines for this reason, offering insights into their unique perceptions related to LA. 
Moreover, we utilize a clustering approach to uncover the nuances in student LA perspectives, distinguishing four primary clusters: {\itshape Enthusiasts}, {\itshape Realists}, {\itshape Cautious} and {\itshape Indifferents}, which we elaborate further. Although \cite{WhitelockWainwright2021} identified groups in the SELAQ patterns, our approach provides deeper insights into specific student concerns. To the best of our knowledge, this is a novel approach in this context.

\section{Study Design} 
\label{sec:Study:Design}
To outline our research approach, we first delve into the context in which the study was carried out.

Our survey was conducted at a university in Fall 2022 and Summer 2023. Students from various faculties and degree courses, such as Computer Science (CS), Architecture (AR), Civil Engineering (CE), Electro-Mechanical Engineering (EM), Sustainability (SU), Surveying (SV), Business Studies (BS) and others (OE) representing additional engineering disciplines, participated in the survey. The short forms in brackets are used throughout this paper as well as in figures and tables. Therefore our data set primarily focuses on the STEM disciplines (Science, Technology, Engineering, and Mathematics), reflecting the core academic offerings of our university of applied sciences. However, it also encompasses areas of sustainability and architecture, integrating elements from arts and humanities. While these inclusions provide a broader perspective, it's important to note that the emphasis and analytical depth may vary compared to the STEM subjects. Therefore, while our data set does include these additional fields, its primary strength and focus remain rooted in the STEM areas.

It is worth to note that the Sustainability program at this university is distinctive. Unlike some programs that may focus solely on social or engineering aspects of Sustainability, this university's program offers an interdisciplinary approach blending rigorous engineering concepts with substantial contributions from humanities. This blend provides a holistic perspective on sustainability, ensuring that graduates are well equipped to tackle complex challenges in this domain. While the Architecture course primarily leans toward the arts, disciplines like Computer Science, Engineering and Business Studies have their distinct orientations. 
Interestingly, despite the relatively small size of the Sustainability student group, we have chosen to examine them separately due to their unique position within the predominantly engineering-focused landscape of the university. Consequently, it's crucial to recognize that the statistical uncertainties for this smaller group (Sustainability) are inherently larger than for the more prominent groups like Civil Engineering. 

\begin{figure}[h]
\begin{centering}
\includegraphics[width=0.97\columnwidth]{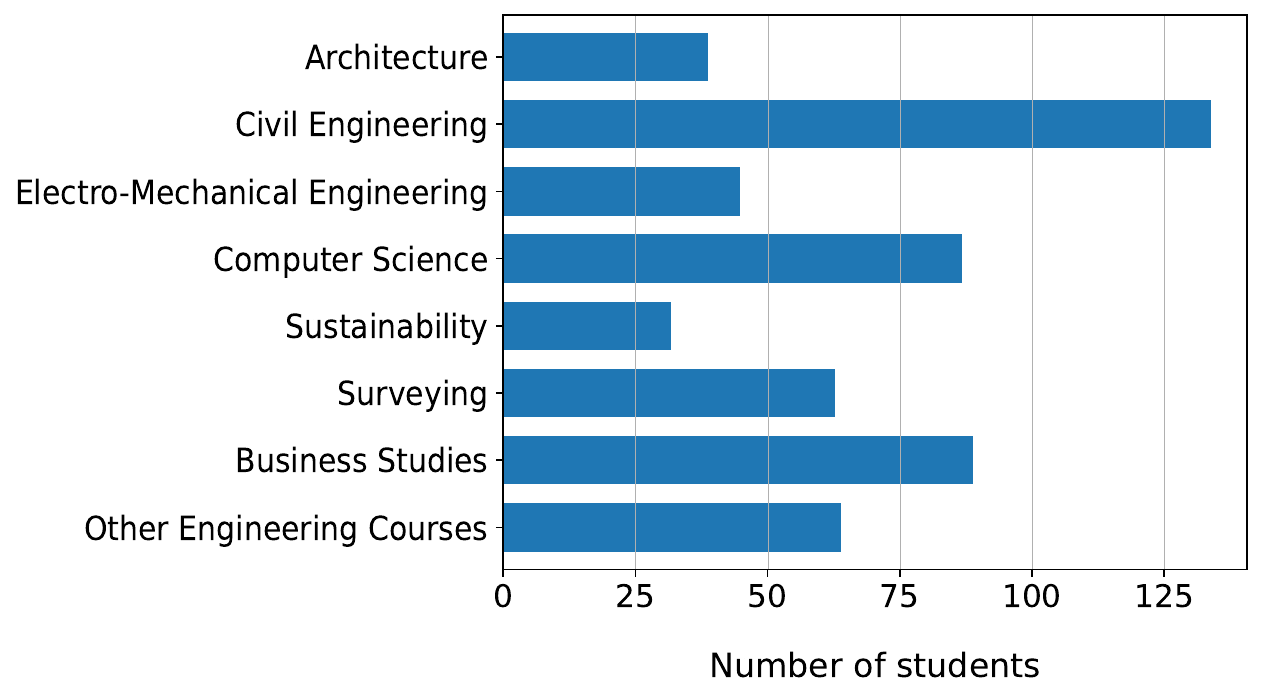}
\par\end{centering}
\caption{Number of students participating in the survey per field of study. Civil engineering, Business Studies and Computer Science are the primary groups.}
\label{fig:stud_pro_fac}
\end{figure}

The SELAQ questionnaire~\cite{whitelock2019student} aims at identifying students' \textit{desires} and \textit{expectations} related to Data Protection within the university context, alongside the utilization of their data for the purpose of LA. This questionnaire was extended with gathering data about degree programs. Additionally, the questionnaire was translated into the official language of the university. Fig.~\ref{fig:stud_pro_fac} illustrates the student numbers by faculties. The survey was conducted in class, ensuring a random sample from these faculties.
The students were presented with 12 questions, for each of which they were required to indicate their desires and expectations through two subquestions:

\begin{itemize}
\item[(d)]  Ideally, I would like that to happen (Desire).
\item[(e)]  In reality, I would expect that to happen (Expectation).
\end{itemize}
To convey their desires (d) and expectations (e) for each of these $12$ questions, students were asked to assign a rating between $1$ (strongly disagree) and $7$ (strongly agree).
The $12$ questions are as follows: 
\begin{enumerate}
\item The university will ask for my consent before using any identifiable
data about myself.
\item The university will ensure that all my educational data will be kept
securely.
\item The university will ask for my consent before my educational data
are outsourced for analysis by third-party companies.
\item The university will regularly update me about my learning progress
based on the analysis of my educational data.
\item The university will ask for my consent to collect, use and analyze
any of my educational data.
\item The university will request further consent if my educational data
are being used for a purpose different to what was originally stated.
\item The Learning Analytics service will be used to promote student decision
making.
\item The Learning Analytics service will show how my learning progress
compares to my learning goals/the course objectives.
\item The Learning Analytics service will present me with a complete profile
of my learning across every module.
\item The teaching staff will be competent in incorporating analytics into
the feedback \& support they provide to me.
\item The teaching staff will have an obligation to act (i.e., support me)
if the analytics show that I am at risk of failing and underperforming
or if I could improve my learning.
\item The feedback from the Learning Analytics service will be used to promote
academic and professional skill development (e.g., essay writing and
referencing) for my future employability.
\end{enumerate}

\begin{figure}[h]
\begin{centering}
\includegraphics[width=0.92\columnwidth]{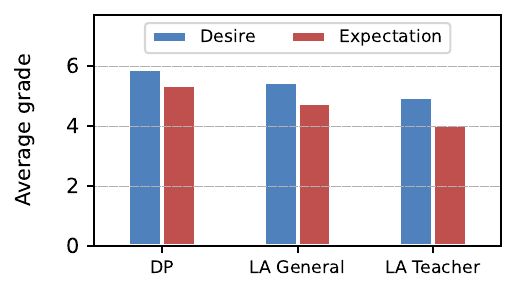}
\par\end{centering}
\caption{Desire (blue) and expectation (red) average grade per question group. For each question group, the expectation is always lower than the desire.}
\label{fig:avg_features}
\end{figure}

The above-mentioned 12 questions encompass the following aspects: Data Protection (DP) features and Learning Analytics features; the latter are divided into LA General and LA Lecturer. Table \ref{tab:SELAQ_question_subgroups} illustrates the alignment of the questions with their respective group showing that 5 questions belong to DP and the remaining 7 to LA.%
\begin{table}[h]
\caption{SELAQ question subgroups}
\centering
\begin{tabular}{|c|c|}
\hline
\textbf{Question Group}     & \textbf{Question Pairs} \\ \hline
Data Protection (DP)            & 1-3, 5, 6               \\ \hline
LA General Functionality (LA General)   & 4, 7-9, 12              \\ \hline
LA Lecturer-related Features (LA Lecturer) & 10, 11                  \\ \hline
\end{tabular}
\label{tab:SELAQ_question_subgroups}
\end{table}%
\noindent A first insight into the study can be gained by summarizing feature averages for the question groups (defined in Table \ref{tab:SELAQ_question_subgroups}), which is shown in Fig.~\ref{fig:avg_features}.
Students consistently expressed not only high expectations regarding the implementation of Data Protection in Learning Analytics but also a strong confidence that the university will uphold these standards as shown in the first two columns.  
The general expectation and desire regarding LA are nearly as high as those for DP. However, when shining a spotlight on the questions related to the lecturer, an intriguing pattern emerges. While the overall expectation about LA remains high, both the anticipation and the confidence that these anticipations will be met decrease noticeably in the context of the lecturers. The effect is even stronger regarding the anticipations.

\section{Analysing the Questionnaire Results}
\label{sec:Study:Analysing:the:Questionaire:Results}
\begin{figure*}[t]
\centering{}
\includegraphics[width=0.9\textwidth]{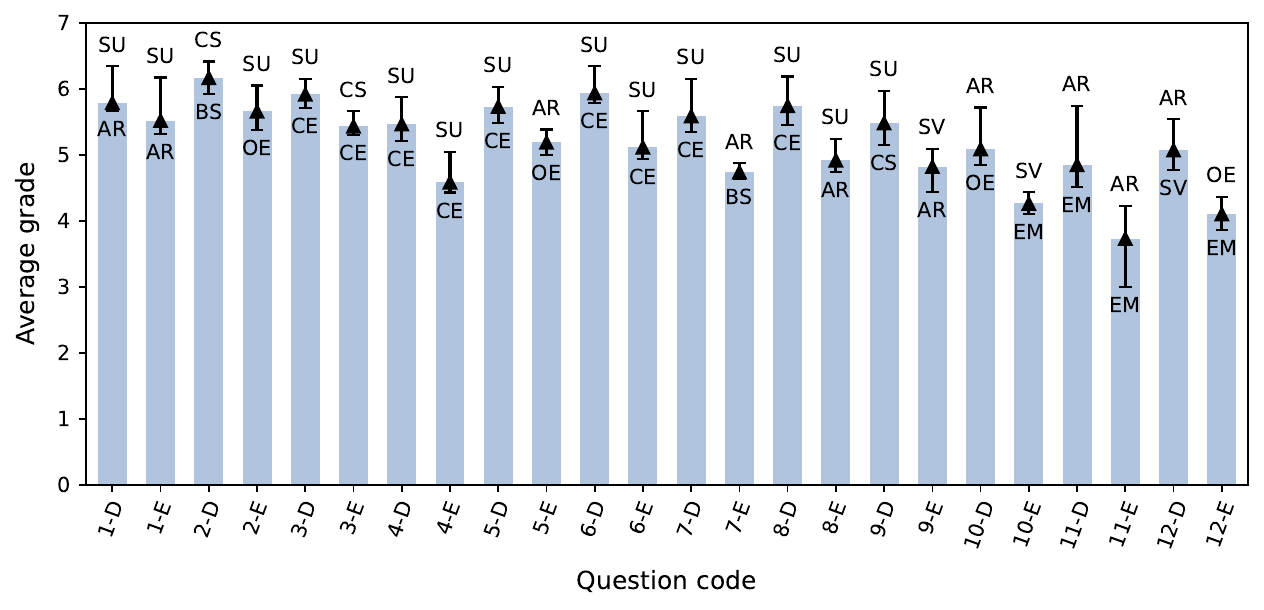}
\caption{Overall average grades by question (bars) and in what degree course (annotations) the lowest / highest average (lines) is measured: Architecture (AR), Civil Engineering (CE), Electro-Mechanical Engineering (EM), Computer Science (CS), Sustainability (SU), Surveying (SV), Business Studies (BS), Other Engineering Courses (OE).}
\label{fig:grade_averages}
\end{figure*}
We aim to analyze the perception, criticism and acceptance of Learning Analytics within the student population. In a first step  missing values were imputed~\cite{Donders2006} with the mean value strategy for numerical features and with the most common value for categorical features. An examination of the raw data suggests that we should compute the average scores for all survey participants. Overall, the average grades are quite high. It is particularly striking that the expectation for question 11 ("The teaching staff will have an obligation to act (i.e., support me) if the analytics show that I am at risk of failing and underperforming or if I could improve my learning.") is significantly lower than for all other questions. It can be deduced that students are not very convinced that their teachers provide support and optimize learning success.
Furthermore, Fig.~\ref{fig:grade_averages} shows that Sustainability students give significantly higher scores in most questions, indicating that these students generally have high desires and expectations.
In contrast, a relatively large number of Civil Engineering students tend to give ratings below the average.
Computer science students express a stronger agreement with statements 2d, 3e and 9d, emphasizing the secure handling of their educational data, consent for third-party analysis, and the desire for a comprehensive learning profile, respectively. This heightened awareness and expectation might be attributed to their specialized background in information technology, where understanding and valuing data privacy and security are paramount. 
These findings suggest that it is worth to check whether the students are homogeneous or whether groups can be formed.
To do so, we employ the K-means clustering algorithm \cite{Hartigan1979} on the responses to the $24$ questions provided by $553$ students. Clustering is a collection of machine learning techniques that combines data into groups 
(called clusters) during unsupervised learning. Utilizing the elbow \cite{Umargono2020} and silhouette \cite{Wang2017} methods, we estimate the optimal value of k, which indicates the number of clusters. In both methods, the analysis indicates that $k$ should be set to $4$. After determining the optimal k, we subsequently rerun the K-means algorithm.

For each cluster we computed average grades of each question group. The outcome shown in Table \ref{tab:average_for_la_dp} allows us to get a deeper understanding on the 4 clusters. 

\begin{table}[H]
\caption{Average students grade for Data Protection (DP) and Learning Analytics (LA) question groups. (d) and (e) stand for desire and expectation, respectively.}
\centering
\begin{tabular}{|l|c|c|c|c|}
\hline
  & \bfseries DP (d)   & \bfseries DP (e)  & \bfseries   LA (d)    & \bfseries LA (e)  \\ \hline
All clusters & 5.91 & 5.39 & 5.33 & 4.51    \\ \hline
Cluster A (\textit{Enthusiasts}) & 6.34 & 6.14 & 6.01 & 5.64   \\ \hline
Cluster B (\textit{Realists}) &  6.34 & 4.95 & 5.76 & 3.73    \\ \hline
Cluster C (\textit{Cautious}) & 6.19 & 5.64 & 3.68 & 3.75     \\ \hline
Cluster D (\textit{Indifferent}) & 3.55 & 4.02 & 4.24 & 3.91      \\ \hline
\end{tabular}
\label{tab:average_for_la_dp}
\end{table}
Students within cluster A exhibit a strong inclination to embrace Learning Analytics with an average rating of $6.01$, perceiving it as a valuable enhancer of their learning journey. They also hold expectations of the university's commitment to implementing LA as promised, marked by an average rating of $5.64$. Although they express a high desire for Data Protection (average rating of $6.34$), they also maintain an expectation of alignment with the university's pledges (average rating of $6.14$). We categorize this cluster as the \textit{Enthusiasts}. 
Students in Cluster B, much like those in Cluster A, exhibit high values in terms of their desire for LA and DP implementation (average rating of $5.76$ and $6.34$, respectively). However, a key distinction arises in their expectations regarding realization. While they hold reservations concerning the actual implementation of DP and LA (average rating of $3.73$ and $4.95$), they remain receptive to the potential benefits it could bring. Hence, we characterize them as \textit{Realists}.
Cluster C stands out with the lowest desire and expectation levels concerning Learning Analytics ($3.68$ and $3.75$). On the contrary, they exhibit robust desires and expectations for Data Protection ($6.19$ and $5.64$). This cohort of students holds reservations regarding the effectiveness of LA and might necessitate greater persuasion and reassurance concerning its implementations. On the other hand, their inclination toward desiring Data Protection and expecting it to be realized remains high. We characterize them as the \textit{Cautious}.

\begin{figure*}[bt]
\begin{centering}
\includegraphics[width=1.5\columnwidth]{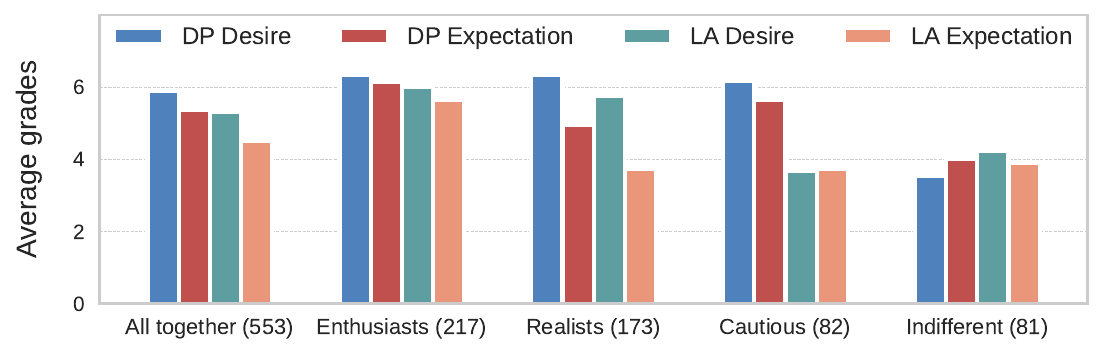}
\par\end{centering}
\caption{Average rating of all respondents and divided according to the clusters formed (with cluster size). Each for both sub-groups (LA, DP) and sub-questions (desire, expectation).}
\label{fig:KMeans_clusters}
\end{figure*}

Fig.~\ref{fig:KMeans_clusters} provides a comprehensive summary of the levels of desire and expectation
\noindent
Students in Cluster D exhibit a general lack of interest or involvement in desire and expectation of both Data Protection and Learning Analytics, showing relatively similar medium average ratings. We designate them as the \textit{Indifferent}. 
within each cluster, along with the designated nomenclature that we have attributed to each of the cluster.
\section{Understanding the Student Clusters}
The aim of this section is to explain in detail the clusters that have been formed and to develop from this an interpretable model for classifying students.

\begin{figure*}[tbh]
    \centering
    \subfigure[Transparent and explainable formation of student clusters through a simple but effective decision tree approach.]
    {
        \includegraphics[width=0.62\columnwidth]{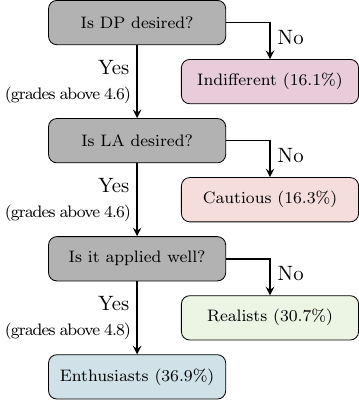}
        \label{fig:decision_tree_clust}
    }
    \hskip 4em\subfigure[Distribution of student attitudes across different fields of study.]
    {
        \includegraphics[width=1.26\columnwidth]{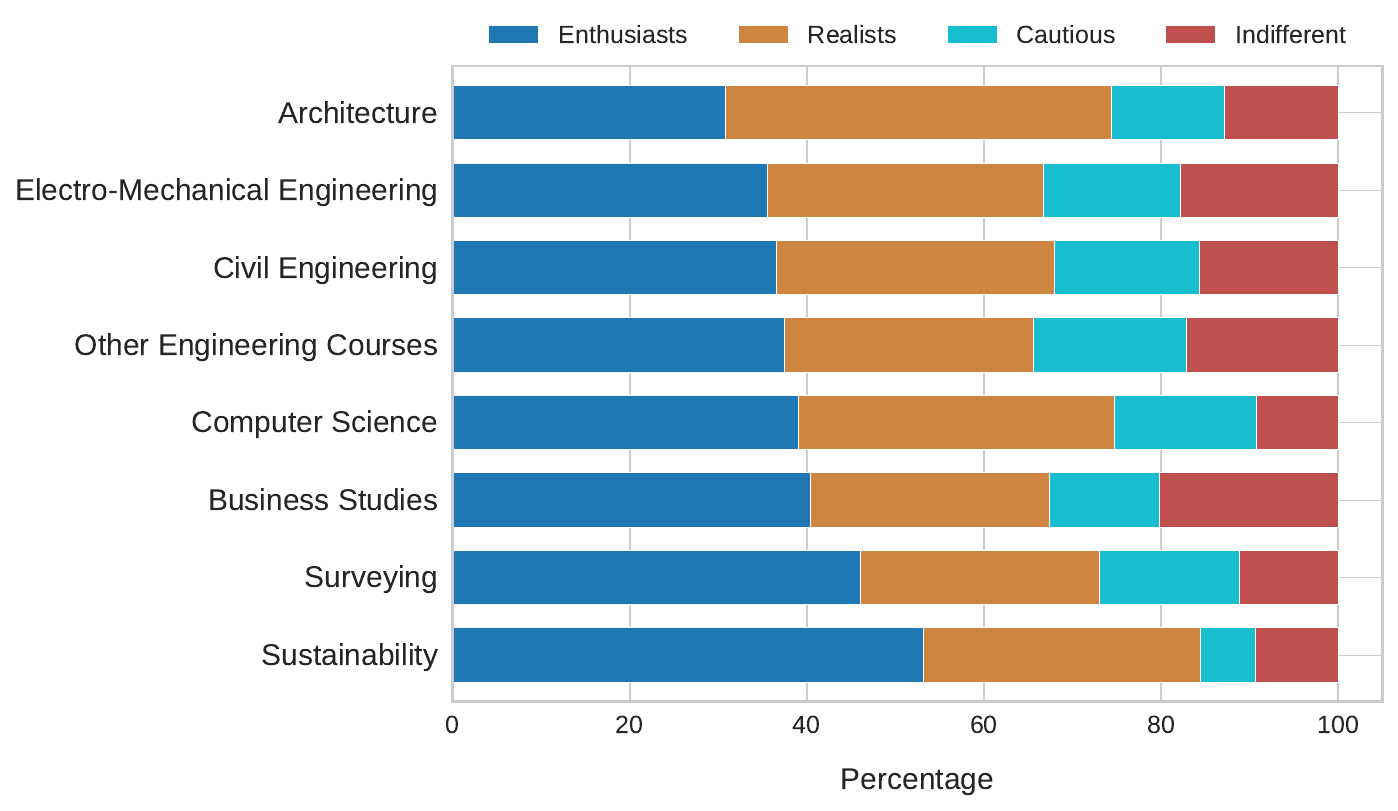}
        \label{fig:cluster_vs_degreecourses}
    }
    \caption{Analyzing Student Clusters and Attitudes: Decision Tree Analysis and Distribution Across Study Fields.}
\end{figure*}

As a next step, we aim to delve deeper into our student cluster findings and validate the characterization of each group, as elucidated in the preceding section. To achieve this, we employ a decision tree methodology \cite{Quinlan1986} to model the clusters seen in Table \ref{tab:average_for_la_dp}. Decision trees are particularly well-suited when prioritizing transparency and explainability \cite{Guidotti2018}. In our analysis, we employ the CART (Classification and Regression Trees) method to learn a decision tree. We use DP and LA as features to classify the students. This approach significantly enhances our understanding of student behaviors and preferences by providing a detailed and comprehensive perspective. During our experiments, we discovered that the rate/grade $4.6$ respective $4.8$ are effective split points (split threshold). The decision tree used this threshold to optimally separate the data points into distinct categories by partitioning the features into branches. The structure of the resulting decision tree is shown in 
 Fig.~\ref{fig:decision_tree_clust}.

One might ask whether Computer Science students, given their greater affinity for technology, exhibit more enthusiasm for LA compared to Architecture or Business students. Another interesting question could be whether Sustainability students are more cautious regarding Data Protection implementations than their Engineering counterparts. To gain insight into this, we model the distribution of our identified clusters (\emph{Enthusiast, Realist, Cautious, Indifferent}) across the different fields of study. The results are reported in Fig.~\ref{fig:cluster_vs_degreecourses}.
As illustrated in Fig.~\ref{fig:cluster_vs_degreecourses}, students from Engineering and Computer Science disciplines exhibited a relatively similar pattern in their behavior. Approximately $20\%$ of these students are identified as \emph{Cautious}, while a notable $40\%$ align with the \emph{Enthusiasts} category. The number of \emph{Indifferents} is smaller for Computer Science students and the number of \emph{Realists} is a bit larger, suggesting that they may have a more differentiated opinion due to their IT affinity. 
On the contrary, students in the field of Architecture exhibit a stronger inclination toward realism, with a larger proportion leaning towards the {\itshape Realist} attitude.
Within the group of Business Administration students, a noteworthy number is categorized in the \emph{Indifferent} category. This suggests a potential sense of overall ambivalence or limited intentional investment in relation to Learning Analytics.
Meanwhile, Sustainability students appear to be more receptive and positive toward LA, as indicated by a higher percentage falling under the {\itshape Enthusiasts} category.
Overall, this distribution suggests that students' attitudes toward LA can significantly vary depending on their field of study. Engineering and Computer Science students, potentially due to their technical background, find themselves at crossroads between enthusiasm and realism concerning the potential advantages of LA, alongside the implementation of DP.
In contrast, Architecture students, with their design-oriented curriculum, might be more critical toward the applicability or relevance of LA within their field. The indifference of Business students could potentially arise from limited exposure or understanding of LA's potential benefits, while Sustainability students, driven by their future-focused outlook, may recognize the potential of utilizing LA for advancing educational prospects.

\section{Discussion}
The results of this study highlight the importance of understanding students' attitudes and expectations of Learning Analytics at universities.
Evidently students' acceptance of LA varies depending on their degree course. Engineering and Computer Science students (technical background), showed a mix of enthusiasm and realism towards LA, likely due to their understanding of the potential benefits and the challenges of DP. On the other hand, Architecture students exhibited a more skeptical attitude, possibly due to the perceived lack of relevance or applicability of LA in their field. Business students displayed a sense of ambivalence or limited investment in LA, suggesting a need for greater exposure and understanding of its potential benefits. Notably, students from non-engineering disciplines, such as Sustainability, Surveying and Business, showed higher levels of enthusiasm towards LA, indicating a potential for broader adoption and integration across various academic disciplines.
Understanding students perspectives within different disciplines can contribute to more effective strategies for a successful implementation of Learning Analytics. 
Another important observation is the small proportion of students categorized as {\itshape Cautious} or {\itshape Indifferent} towards LA. While these clusters represent a minority, they require special attention and consideration. There can be multiple reasons why students express indifference towards LA and DP. Possible lack of interest or knowledge as well as the perception of insignificance of the topic are linked to the content itself. A general indifference towards study or university related topics may stem from doubts or discontent with studying, lack of identification with the university or personal issues in general. Efforts should be made to address the concerns and reservations of these groups, so universities can ensure a more comprehensive implementation of LA. We suggest that {\itshape Enthusiasts}, who are naturally inclined towards LA and for whom the primary focus should be on guaranteeing the robustness of LA systems and delivering on promised features. So that we do not lose these students. The {\itshape Realists} exhibit skepticism about institutional capabilities in LA; to address this, it is essential to reinforce the institution's digital competence, underscore strict DP standards and foster transparent communication to alleviate their concerns. The {\itshape Cautious} mainly hesitant about LA's effectiveness. Engaging them requires workshops detailing LA's benefits and emphasizing the rigor of DP, highlighting tangible outcomes and security measures. On the other hand, the {\itshape Indifferent} group, characterized by general ambivalence, may resonate with strategies that combine approaches effective for both the {\itshape Realists} and {\itshape Cautious}. Furthermore, a discernible pattern emerges from our data, indicating a reduced trust in lecturers' ability to effectively use LA. Addressing this challenge involves providing specialized LA training for lecturers and promoting open, transparent discussions about LA's utility and role of in the academic process.
Importantly, universities with pronounced academic specializations should tailor their strategies. Evidence suggests that the proportions of these clusters might vary depending on the academic discipline, making it essential for institutions to recognize and adjust based on the dominant student cluster within their milieu. In conclusion, melding a nuanced understanding of student profiles with considerations tailored to specific academic disciplines is pivotal for the successful and smooth integration of Learning Analytics.

\section{Conclusion and Future Prospects}
In this study, we conducted a survey to shed light on the acceptance of Learning Analytics among university students and to highlight the significance of understanding their attitudes and expectations.
After examining the survey results and distinguishing responses among student groups by means of an automatic clustering approach, we clustered four different student groups.
A decision-based approach also revealed certain student attitudes and preferences according to different academic disciplines. These insights give a foundation for successful and broadly accepted implementations of LA at universities. Several questions arise from these findings. Firstly, how can instructors effectively communicate with and address the concerns of cautious students, ensuring their trust in the use of LA? Secondly, what strategies can be employed to enhance the engagement of indifferent students and highlight the advantages that LA can offer? Additionally, students of mainly technical subjects participated in the present study. Future studies could explore whether expectations and concerns about LA of students from other domains (such as Humanities or Social Sciences) differ significantly from our findings.
\begin{acks}
This work is a part of the Digital Mentoring project funded by the Stiftung Innovation in der
Hochschullehre under FBM2020-VA-219-2-05750. Additionally, this work was funded by the BMBF under 16-DHB-4021. The authors would like to express sincere gratitude to their colleague C. Kaufmann for his help.
\end{acks}
\bibliographystyle{ACM-Reference-Format}
\bibliography{references}

\end{document}